\documentclass[conference]{IEEEtran}
\IEEEoverridecommandlockouts

\usepackage{cite}
\usepackage{amsmath,amssymb,amsfonts}
\usepackage{algorithmic}
\usepackage{graphicx}
\usepackage{textcomp}
\usepackage{xcolor}

\def\BibTeX{{\rm B\kern-.05em{\sc i\kern-.025em b}\kern-.08em
    T\kern-.1667em\lower.7ex\hbox{E}\kern-.125emX}}
\begin{document}

\title{The Use of Video Captioning for Fostering Physical Activity\\

}
\makeatletter
\newcommand{\linebreakand}{%
  \end{@IEEEauthorhalign}
  \hfill\mbox{}\par
  \mbox{}\hfill\begin{@IEEEauthorhalign}
}
\makeatother

\author{\IEEEauthorblockN{Soheyla Amirian}
\IEEEauthorblockA{\textit{dept. Computer Science} \\
\textit{The University of Georgia}\\
Athens, GA, USA \\
amirian@uga.edu}
\and
\IEEEauthorblockN{Abolfazl Farahani}
\IEEEauthorblockA{\textit{dept. Computer Science} \\
\textit{The University of Georgia}\\
Athens, GA, USA \\
a.farahani@uga.edu}
\and
\IEEEauthorblockN{Hamid R. Arabnia}
\IEEEauthorblockA{\textit{dept. Computer Science} \\
\textit{The University of Georgia)}\\
Athens, GA, USA \\
hra@uga.edu}
\linebreakand
\IEEEauthorblockN{Khaled Rasheed}
\IEEEauthorblockA{\textit{dept. Computer Science} \\
\textit{The University of Georgia}\\
Athens, GA, USA \\
khaled@uga.edu}
\and
\IEEEauthorblockN{Thiab R. Taha}
\IEEEauthorblockA{\textit{dept. Computer Science} \\
\textit{The University of Georgia}\\
Athens, GA, USA \\
trtaha@uga.edu}
}

\maketitle
\thispagestyle{plain}
\pagestyle{plain}

\begin{abstract}
Video Captioning is considered to be one of the most challenging problems in the field of computer vision. Video Captioning involves the combination of different deep learning models to perform object detection, action detection, and localization by processing a sequence of image frames. It is crucial to consider the sequence of actions in a video in order to generate a meaningful description of the overall action event.
A reliable, accurate, and real-time video captioning method can be used in many applications. However, this paper focuses on one application: video captioning for fostering and facilitating physical activities. In broad terms, the work can be considered to be assistive technology.
Lack of physical activity appears to be increasingly widespread in many nations due to many factors, the most important being the convenience that technology has provided in workplaces. The adopted sedentary lifestyle is becoming a significant public health issue. Therefore, it is essential to incorporate more physical movements into our daily lives.
Tracking one's daily physical activities would offer a base for comparison with activities performed in subsequent days. With the above in mind, this paper proposes a video captioning framework that aims to describe the activities in a video and estimate a person's daily physical activity level. This framework could potentially help people trace their daily movements to reduce an inactive lifestyle's health risks. The work presented in this paper is still in its infancy. The initial steps of the application are outlined in this paper. Based on our preliminary research, this project has great merit.

\end{abstract}
\begin{IEEEkeywords}
Deep Learning, Video Captioning, Object Detection, Recurrent Neural Network, Natural Language Processing.
\end{IEEEkeywords}

\section{Introduction}
\label{sec:intro}
With the recent availability of powerful machines (GPUs, CPU clusters), together with large amounts of training data, deep learning has made a come back providing breakthroughs on image recognition, and object detection \cite{he2016deep, toutiaee2020video, amirian2018,shenavarmasouleh2020drdr, shenavarmasouleh2020drdr2,9281287} with many applications. Object detection models \cite{he2016deep, redmon2017yolo9000} are used to extract object information and localization from images and videos.
The prosperity of object detection models has made them suitable for other deep learning tasks, including image and video captioning, automation of making a title for videos \cite{9281287, title2020} and more.

Image captioning is a comparatively more trivial task than video captioning as we have to deal with more objects in a video. Video captioning requires understanding the video contents accurately to detect the objects, their corresponding actions, and relations.  In this task, detecting the objects' correlation plays an important role in generating a meaningful and consistent description. Besides, events may overlap as the length of events varies across videos, making object detection a difficult task. Some existing models address these concerns \cite{9281287, krishna2017dense, mun2019streamlined,aafaq2019spatio}.
A reliable, accurate, and real-time video captioning method can be used in many applications.
Video captioning techniques are utilized in
medical and healthcare applications,
a guide to interacting with people with visual impairments,
human-robot interaction,
automatic video subtitling,
video surveillance,
automatic title/summary generation \cite{title2020},
self-driving vehicles, sign language translation, and many others \cite{9281287}. Technology is moving faster than ever, and we could soon interact with robots in the same manner as we do with humans. Taking advantage of human-robot interaction, medical and healthcare applications, we introduce a novel video captioning framework that offers an activity summary.
By using video captioning for fostering and facilitating physical activities, the proposed model can be used as assistive technology.

Lack of physical activity is increasingly widespread in many nations due to many reasons. The adopted sedentary lifestyle is becoming a significant public health issue. Therefore, it is essential to incorporate more physical movements into our daily lives.
Tracking one's daily physical activities would offer a base for comparison with activities performed in subsequent days. With the above in mind, this study proposes a video captioning framework that aims to describe the activities in a video and estimate a person's daily physical activity level. This framework could potentially help people trace their daily movements to reduce the health risks of an inactive lifestyle.
First, we feed a video of a person's daily activity into the model. Then, model processes the video by extracting actions, and generating captions.  Finally, it summarizes an activity history. Figure \ref{fig:process} illustrate more details.
This framework could potentially help people trace their daily movements to reduce the health risks of an inactive lifestyle by managing their activities \cite{farahani2020brief}.

The main contribution of this research is proposing a novel video captioning framework. This framework utilizes the Spatio-Temporal information in a video to generate accurate and coherent captions for filmed physical activities. The captions comprise temporal dynamics of discovered actions that could be used to follow a person's physical activity in daily life.
The work presented in this paper is still in its infancy. The initial steps of the application are outlined in this paper. Based on our preliminary research, this project could be a healthcare application used by physicians or the public.

\section{Literature Review}
\label{sec:related}
\begin{figure*}[!ht]
    \centering
    \includegraphics[width=\linewidth]{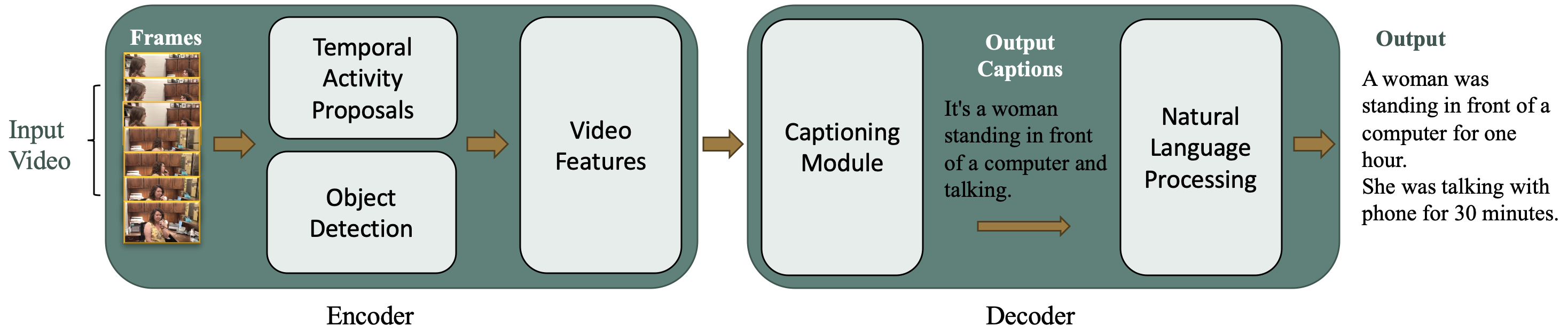}
    \caption{This is an overview of the proposed framework. It consists of an encoder and a decoder. The encoder part focuses on detecting objects, multiple events, and actions recorded in a video by jointly localizing temporal proposals of interest. The decoder first generates captions for each event proposal and then finds the correlation between them to summarize an activity history. The produced story describes the type of each physical activity and the corresponding duration. We can use this information to categorize the video into different physical activity levels.}
    \label{fig:process}
\end{figure*}
\subsection{Video Captioning.}
Video captioning is an automatic process that aims to generate natural language sentences describing a given video's contents. Video captioning can be considered a system describing human activities with natural language \cite{9281287}. Recent video captioning techniques often consist of two main parts; an encoder and a decoder.
In Dense-captioning proposed by Krishna et al. \cite{krishna2017dense}, the encoder part focuses on detecting multiple events recorded in a video by jointly localizing temporal proposals of interest, and the decoder describes the events in natural language sentences. This model introduces a new captioning module that applies the past and future contextual information to describe all the events jointly.
Mun et al. \cite{mun2019streamlined} extend the Dense-captioning model by proposing Streamlined Dense Video Captioning. This technique utilizes multiple steps to generate coherent, consistent, and unique video descriptions by incorporating temporal dependencies across events. To achieve this, the model first introduces the Event Proposal Network (EPN) that adopts Single-Stream Temporal action proposals (SST)\cite{buch2017sst}. EPN tends to find a series of candidate event proposals that are semantically and temporally meaningful. In the next step, the Event Sequence Generation Network (ESGN) creates an episode for a video by selecting a highly correlated set of event proposals from the candidate proposals. Finally, SCN generates coherent captions for the selected event proposals using a sequential captioning network (SCN). SCN consists of an episode RNN and an event RNN. The episode RNN takes the selected proposals one at a time to model an episode's state, and the event RNN generates the caption for each event proposal in a sequential manner, where the words in a caption are conditioned on the implicit representation of the episode.

\subsection{Temporal Activity Proposals.}
Action recognition is an essential part of video captioning. This task generally provides fundamental tools and applications for action detection. For example, Temporal Action Detection is a tool that focuses on localizing the temporal extent of each action for a detected object \cite{caba2016fast}. 
Heilbron et al. \cite{caba2016fast} proposed Fast temporal activity proposals to detect human actions in untrimmed videos efficiently. This method is an end-to-end action detection pipeline that generates high-quality proposals in terms of localization and ranking. However, the proposal method's efficiency needs to be improved by interleaving or combining the feature extraction and proposal representation step.
SST \cite{buch2017sst} can run continuously in a single stream over long input video sequences to find semantically meaningful temporal regions via a single scan of videos.
SST produces high temporal overlapping proposals with grand truth action intervals by considering and evaluating many action proposals over densely sampled time-scales and locations. In this technique, \textit{Input} is a video consist of \textit{N} frames, and \textit{Visual Encoding} applies 3D Convolutional (C3D) network \cite{tran2015learning} to compute feature representation. \textit{Sequence Encoding}
progressively accumulates the evidence in a video sequence and simultaneously disregards the irrelevant background. This process continues until it is confident that an action is taking place in the video.
Finally,  \textit{Output} produces confidence scores of multiple proposals at each time step. 3D-CNNs are powerful tools to learn Spatio-temporal features by encapsulating information related to objects, scenes, and actions. They are popular networks, and many recent video captioning frameworks \cite{mun2019streamlined, aafaq2019spatio, wang2018bidirectional} benefit from them.
Yang et al. \cite{yang2019step} proposed the first end-to-end progressive optimization framework for video action detection known as Spatio-Temporal Progressive Learning (STEP). STEP recognizes the action of interset recorded in a video by localizing them in both space and time. Unlike previous methods that directly perform action detection in one run, STEP involves a multi-step optimization process that progressively refines the initial proposals towards the final solution with spatial refinement and temporal extension. Multiple steps run in a sequential order, where the outputs of one step are used as the proposals for the next step. Spatial refinement aims to improve the classification and localization of the action's regions. It starts with a small number of coarse-scale proposals and updates them iteratively. Temporal extension, on the other hand, focuses on improving classification accuracy by incorporating longer-range temporal information.
Moreover, STEP achieves superior performance by using only a handful of proposals and obviating the need to generate and process large numbers of proposals.

\subsection{Object Detection.}
Object detection is a computer vision task that tends to collect the related tasks to identify objects in an image or video frame by localizing and classifying them accurately. Deep learning models \cite{amirian2018} have helped object detection by extracting high-quality representations that result in identifying more objects with precise localization. Many video captioning techniques benefit from object detection models to improve captioning performance.
Redmon et al. \cite{redmon2017yolo9000} proposed YOLO (You Only Look Once), a real-time object detection system that can detect over 9000 object categories. Aafaq et al. \cite{aafaq2019spatio} utilized YOLO in their video captioning framework. The proposed model processes the objects' locations and the corresponding multiplicity information extracted from YOLO to encode the scenes' spatial dynamics.

\subsection{Decoder.}
Hierarchical recurrent neural networks are employed to generate coherent captions based on the detected events. Mun et al. \cite{mun2019streamlined} proposed framework includes a captioning network termed as Sequential Captioning Network (SCN) that generates the descriptions for the event proposals provided by other parts of the framework. SCN consists of two parts; episode RNN and event RNN. Episode RNN adopts a single-layer Long Short Term Memory (LSTM) with a 512-dimensional hidden state while event RNN utilizes captioning network with temporal dynamic attention and context gating (TDA-CG) \cite{wang2018bidirectional}. Reinforcement Learning is applied to learn RNNs in SCN using event and episode-level rewards. Event-level reward allows the system to accurately capture specific content in each event while the episode-level reward enforces the network to produce coherent descriptions from all generated captions.
Dense-Captioning \cite{krishna2017dense} uses a captioning LSTM network to describe the video. It develops an analogous model that groups the events to capture the temporal context. The captioning module of the framework then incorporates the context from neighboring events to capture the correlations between events.
Instead of using a complex language model, Aafaq et al. \cite{aafaq2019spatio} utilized multiple layers of Gated Recurrent Units (GRUs) \cite{cho2014learning} in the captioning part of their model. Gated Recurrent Units are simple sequential model and known to be robust to vanishing gradient problem. GRUs generate semantically rich captions from the visual representation discovered by the earlier part of the model.

\section{Proposed Framework}
\label{sec:method}

In this section, we elaborate on the proposed idea for a video captioning framework. The model intends to generate a meaningful, coherent, and accurate video description for the overall action events by incorporating the Spatio-Temporal information of each event recorded in the video. 
The proposed technique involves two main components; an Encoder and a Decoder. The encoder utilizes multiple deep learning models to discover high-quality representations from the input video frames. These representations are then embedded into a high-dimensional feature space to create an input for the decoder. The decoder comprises a captioning module and a Natural Language Processing (NLP) module. The captioning module generates a description for each input video frame, while the NLP module aims to create a coherent story for all input videos by considering the generated captions and finding the correlation between them. In the following, we describe our framework in detail.
\subsection{Encoder}
In the encoding section, we propose a visual encoding technique for video action detection. This technique aims to compute representations enriched by utilizing Spatio-TEmporal Progressive (STEP) action detector \cite{yang2019step} to recognize the actions of interest presented in a video while considering their spatial and temporal properties. STEP consists of spatial refinement and temporal extension, where the former tends to update the accurate localizations for the proposals iteratively, and the latter gradually incorporate relevant temporal context by increasing sequence length. 
STEP starts with a small number of initial proposals and updates them to better classify and localize action regions in the spatial refinement. In this section, multiple steps are carried out in a sequential order, where the outputs of one step are utilized as the proposals for the next step. Temporal extension, on the other hand, improves the classification accuracy by including longer-range temporal information.
Generally, we aim to embed object labels, their frequencies of occurrence, and the evolution of their spatial locations in the encoding process. In this pipeline, the temporal information is kept to be further utilized in the framework's final output. The decoding part of the framework uses this information to keep track of each action's duration.
Furthermore, we encode spatial dynamics by processing objects' locations and their multiplicity information extracted from an Object Detector (i.e., YOLO \cite{redmon2017yolo9000, redmon2018yolov3}). The output layers in both Object Detector and STEP are then fused to extract the object's semantics and actions of the video.
\subsection{Decoder}
The decoding section processes the representations extracted by the encoder to generate the captions describing each action event recorded in the video. Captioning can be obtained by naively treating each action individually. However, events are usually highly correlated in a video, and ignoring the correlations between actions could generate inconsistent or redundant descriptions. To address this challenge, the model needs to include the temporal dependency across events. Hence, our framework's decoder adopts the sequence modeling components of \cite{aafaq2019spatio}, which employs Gated Recurrent Units (GRU).
We specifically employ two layers of GRUs in our model to use the temporal information produced by the encoder. Moreover, we utilize Natural Language Processing system \cite{devlin2019bert} to find the relevant actions and calculate the duration for each action by analyzing the descriptions generated by GRUs. It helps the model to automate the process of generating the final report. The overall framework of the proposed method is illustrated in Figure \ref{fig:process}.

\section{Discussion and Future Work}
\label{sec:conclusion}

Deep learning and video captioning frameworks can be used as technologies to help people to control their health.
The captions can be used as meta-data for search engines, taking the search engine's functionality to a new dimension. Captions can also be used as part of recommendation systems in many applications.
Video captioning is a very challenging task as it is naturally very compute-intensive. The currently proposed captioning models can only deal with short videos that are a few seconds long. 
With the next generation of GPUs and explicit parallel algorithms targeted at the GPU machine architectures, we could achieve real-time performance for longer videos. 
A great opportunity in the area of video captioning is to design and develop a strategy that would permit users to request video captions with various levels of details.
In this research, we investigated different frameworks of video captioning and action detection models. We proposed a framework that takes a video from users containing their daily activity. Then, it analyzes and captions the video to inform the user about their physical activity in daily life.
We are currently implementing this framework that could potentially be used as a health recommendation application.
Moreover, we are not concerned about the efficiency of the framework at the execution time. We hope to be able to address execution efficiency issues in our subsequent publications. 

\section{Acknowledgement}
\label{sec:ack}
We gratefully acknowledge the support of NVIDIA Corporation with the donation of the Titan V GPU used for this research.

\bibliographystyle{IEEEtran}

\end{document}